\documentclass[10pt,twocolumn,letterpaper]{article}

\usepackage{iccv}
\makeatletter
\@namedef{ver@everyshi.sty}{}
\makeatother
\usepackage{tikz}
\usepackage{times}
\usepackage{epsfig}
\usepackage{graphicx}
\usepackage{amsmath}
\usepackage{amssymb}

\usepackage[pagebackref=true,breaklinks=true,letterpaper=true,colorlinks,bookmarks=false]{hyperref}

\usepackage[capitalize]{cleveref}
\crefname{section}{Sec.}{Secs.}
\Crefname{section}{Section}{Sections}
\Crefname{table}{Table}{Tables}
\crefname{table}{Tab.}{Tabs.}

\usepackage[subrefformat=parens]{subcaption}
\usepackage{balance}
\usepackage{cite}
\usepackage{graphicx}
\usepackage{booktabs}
\usepackage{tabularx}
\usepackage{multirow}
\usepackage{multicol}
\usepackage{cite}
\usepackage{bm}
\usepackage{url}
\usepackage{balance}
\usepackage{color}
\usepackage{bbding}
\usepackage{nccmath}
\usepackage{xspace}
\usepackage{wrapfig}
\usepackage{colortbl}
\usepackage{todonotes}
\usepackage[capitalize]{cleveref}
\usepackage{mleftright}
\usepackage{mymacro} 
\crefname{section}{Sec.}{Secs.}
\Crefname{section}{Section}{Sections}
\Crefname{table}{Table}{Tables}
\crefname{table}{Tab.}{Tabs.}

\iccvfinalcopy 

\def\iccvPaperID{4718} 

\ificcvfinal\fi

\begin{document}

\title{InCrowdFormer: On-Ground Pedestrian World Model From Egocentric Views}

\author{Mai Nishimura\textsuperscript{\rm 1,\rm 2}\\
\textsuperscript{\rm 1}OMRON SINIC X\\
Tokyo, Japan\\
{\tt\small mai.nishimura@sinicx.com}
\and
Shohei Nobuhara\textsuperscript{\rm 2}\;\;\;\;\;\;\;\;\;\;\;\;\;\;\;\;\;\;\;\;\;\;
Ko Nishino\textsuperscript{\rm 2}\\
\textsuperscript{\rm 2}Kyoto University\\
Kyoto, Japan\\
{\tt\small \{nob,kon\}@i.kyoto-u.ac.jp}
}

\makeatletter
\let\@oldmaketitle\@maketitle%
\renewcommand{\@maketitle}{
    \@oldmaketitle%
    \vspace{-5mm}
    {
    \centering
        \includegraphics[width=0.95\linewidth]{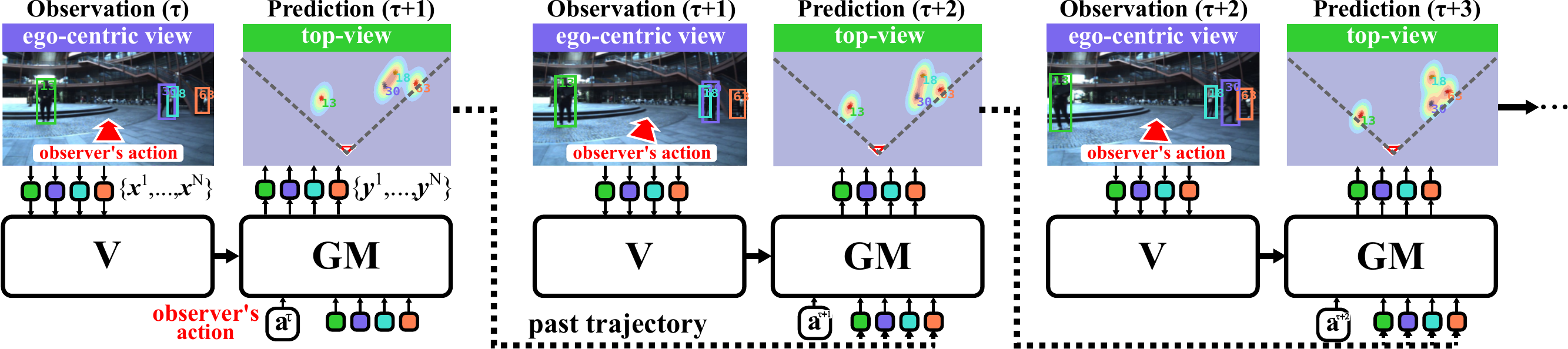}
    \captionof{figure}{We introduce \methodname, the first-of-its-kind Pedestrian World Model that \textbf{predicts} the \textbf{on-ground} future states of pedestrians solely \textbf{from ego-centric} views captured by a \textbf{moving observer}.
    Given in-image $N$ pedestrian states $\{\bm x^1,\dots,\bm x^N\}$ and an observer's action $\bm a_\tau$, \ie, ego-motion, our aim is to predict the on-ground future location of each pedestrian $\{\bm y^1, \dots, \bm y^N\}$.
    \methodname consists of Vision Module (V) and Geometric Memory Module (GM) which use attention for object-centric state encoding and view transformation while considering interactions between the objects.
    } 
    \label{fig:teaser}
    }
    \vspace{4mm}
}
\makeatother

\maketitle

\begin{abstract}
We introduce an on-ground Pedestrian World Model, a computational model that can predict how pedestrians move around an observer in the crowd on the ground plane, but from just the egocentric-views of the observer. Our model, \methodname, fully leverages the Transformer architecture by modeling pedestrian interaction and egocentric to top-down view transformation with attention, and autoregressively predicts on-ground positions of a variable number of people with an encoder-decoder architecture. We encode the uncertainties arising from unknown pedestrian heights with latent codes to predict the posterior distributions of pedestrian positions. We validate the effectiveness of \methodname on a novel prediction benchmark of real movements. The results show that \methodname accurately predicts the future coordination of pedestrians. To the best of our knowledge, \methodname is the first-of-its-kind pedestrian world model which we believe will benefit a wide range of egocentric-view applications including crowd navigation, tracking, and synthesis. 
\end{abstract}

\vspace{-12pt}
\section{Introduction}
\label{sec:intro}
Everyday, we successfully maneuver in a highly dynamic world with just our egocentric, limited view of the surroundings. When we walk down the street, we constantly map the positions of surrounding pedestrians on the ground plane. We not only keep track of their current positions, but also predict their next positions. When we play football, for instance, we are able to tell where other players will end up, which is exactly why we can make that killer pass. 

This predictive mental model of our dynamic surroundings is an illustrative example of ``World Models''~\cite{ha2018worldmodels}, transition models of the environment. In this paper, we are particularly interested in deriving a world model of pedestrians that can continuously localize all visible surrounding people and predict their movements with respect to the known actions of oneself in the next few time steps. Unlike conventional trajectory forecasting methods~\cite{yuan2021agent,alahi2016social,gupta2018social} which require accurate past on-ground pedestrian positions captured by a static surveillance camera, our aim is to predict the on-ground pedestrian positions in the future solely from the in-crowd observer's ego-centric view. The observer is either a person or mobile agent whose movement can be measured (\ie, actions are known).

A world model of pedestrians that can be useful for such downstream tasks, however, requires significant departures from past models~\cite{ha2018worldmodels,kipf2019contrastive,chen2022transdreamer}. First, it must model the pedestrians on the ground plane but from inputs in egocentric perspectives. This requires a nontrivial view transformation, often referred to as Bird's-Eye-View (BEV) transform. The world model, however, needs to predict the next on-ground movements from the observed current movements in the 2D egocentric view, unlike BEV transform that focuses on an image to image (\ie, appearance) transform of only the current frame. We refer to this inherently predictive purely geometric transform as predictive ego2top transform.

Second, the model needs to be fundamentally object-centric. It must model each pedestrian as an independent object that interacts with other objects including the ego-viewer, too. The interaction between pedestrians lies at the heart of the coordination of the pedestrians as a whole, and the transitions of individual pedestrians are largely governed by these object interactions especially in denser crowds. Third, the model needs to model and predict the pedestrian movements and their interactions conditioned on the observer. The observer itself is part of the crowd, and its/his/her actions influence and are affected by the surrounding people. Finally, the viewer would be moving and looking around while walking in the very crowd that needs to be modeled. As a result, surrounding people will come in and out of sight. The crowd itself will also consist of different numbers of people from time to time. The model thus needs to naturally handle a varying number of people. 

We derive a pedestrian world model that satisfies these requirements, namely predictive ego2top transform, object-centric interaction encoding, observer action conditioned modeling, and variable number of constituents. We refer to our model as \emph{\methodname}. The key idea is to model pedestrians as individual tokens and fully leverage attention for modeling their interactions and predictive ego2top transform in a Transformer architecture which also naturally models varying numbers of pedestrians. Unlike past generic world models~\cite{Dugas2022navdreams,chen2022transdreamer} which separate spatial encoding and temporal prediction, our \methodname simultaneously encodes the social, temporal, and geometric relationships of the viewer and surrounding pedestrians.

Two key challenges underlie learning an accurate Pedestrian World Model. The model needs to decouple the ego-motion and pedestrian trajectories from degenerated 2D movements in the image. In addition, the unknown absolute scale of each object, \ie, pedestrian heights, introduces uncertainty in the predictive ego2top transform. For this, we introduce a latent code for each pedestrian and construct a generative model that outputs the on-ground future location distribution conditioned on the observed movements in an ego-centric view and the observer's action. We efficiently model this conditional probability with a Transformer consisting of self-attention that encodes the social and temporal relationship between the observer's action and each pedestrian, and cross-attention that models the geometric relationship between the ego-centric and on-ground views. 

In summary, our contributions are threefold. (1) We introduce, to our knowledge, the first egocentric Pedestrian World Model that models pedestrian and in-crowd observer transitions on the ground plane from egocentric observations, (2) derive it as a novel Transformer that leverages attention for pedestrian interaction modeling and view transform for a variable number of people, and (3) demonstrate its accuracy on real-world crowd motions. Extensive experimental evaluation show that our unified Transformer World Model can accurately predict the future coordination of pedestrians given the observer's action while taking into account uncertainty arising from imperfect cues of depths. We also demonstrate the application of our pretrained model to real video sequences. We believe our \emph{\methodname} will serve as a sound foundation of pedestrian movement modeling for a wide range of applications. We will release our code and data upon acceptance.

\section{Related Work}

\begin{table}[t]
    \centering
    \footnotesize
        \caption{We introduce the first Pedestrian World Model (PWM) that models pedestrian movements solely from their in-crowd egocentric observations with a unified Transformer architecture that embodies predictive ego2top transform and action-conditioned prediction.
        }
\begingroup
\renewcommand{\arraystretch}{0.9}
   \begin{tabularx}{\linewidth}{lccccc}
   \toprule[1.5pt]
    Method & state & ego2top & action &  V & M \\
    \midrule[0.5pt] 
    RSSM~\cite{ha2018worldmodels} & image & -- & \checkmark & CNN & GRU \\
    TSSM~\cite{chen2022transdreamer,Dugas2022navdreams} & image & --  & \checkmark & CNN & Transformer \\
    \midrule[0.5pt]
    C-SWM~\cite{kipf2019contrastive} & object & -- & -- & CNN & GNN \\
    G-SWM~\cite{lin2020improving} & object & -- & -- & CNN & RNN\\
    Slotformer~\cite{wu2022slotformer} & object & -- & --& CNN & Transformer \\
    \midrule[0.5pt]
     \rowcolor[rgb]{0.93,1.0,0.87}\textbf{PWM~(Ours)} & object & \checkmark &  \checkmark & \multicolumn{2}{c}{Transformer} \\
    \bottomrule[1.5pt]
    \end{tabularx}
    \label{tab:wm}
\endgroup
\end{table}

\label{sec:related}

As shown in \Cref{tab:wm}, to the best of our knowledge, we derive the first object-centric world model that encodes pedestrian movements on the ground plane from egocentric views together with the in-crowd observer's actions and its mutual dependency in a unified Transformer model.

\vspace{-12pt}
\paragraph{World Models} A world model is an abstract representation of our environment and its transitions~\cite{johnson1983mental}. Ha and Schmidhuber recently introduced the idea of building a world model with a perception model (V) and a transition model (M), so that a simple agent controller (C)~\cite{ha2018worldmodels} can be learned in the world model with generated rollouts of simulated experiences~\cite{hafner2019learning,hafner2019dream,hafner2021mastering,Dugas2022navdreams}. Most such world models encode the whole image into a latent code as the environment representation. Structured World Models~\cite{kipf2019contrastive,wu2022slotformer} and Object-centric World Models~\cite{lin2020improving} learn object-oriented representations about the world (\ie, individual objects constitute the environment representation). Even in these models, however, the observer never interacts with the environment and object transitions are simply observed from static viewpoints. In sharp contrast, we are interested in modeling a human-populated environment with an object-centric world model that also contains the dynamic observer itself. 

\vspace{-12pt}
\paragraph{Crowd Navigation} 
To plan a feasible path while avoiding collisions with nearby pedestrians, typical approaches predict the future location of pedestrians with analytical models~\cite{helbing1995social,van2011reciprocal} or learning-based crowd models such as social pooling~\cite{chen2019crowd,alahi2016social} and graph neural networks~\cite{chen2020relational,chen2020robot,ivanovic2019trajectron}. 
These methods, however, usually assume direct top-down observations of on-ground pedestrian positions, which is hardly plausible in the real world. 
Few works tackle ego-centric view navigation, \ie, path planning directly from an observer's ego-centric view in the crowd. Dugas \etal~\cite{Dugas2022navdreams} constructed synthetically generated human-populated environments with a game-engine for vision-based navigation. The domain gap between real and synthetic environments is, however, not negligible both in appearances and pedestrian movements. In contrast, our egocentric pedestrian world model is independent of pedestrian appearance and our dataset can be easily augmented with arbitrary combinations of observer actions and real crowd trajectories. 

\vspace{-12pt}
\paragraph{Ego2Top Transformation} Transforming first-person-view (FPV) images into top-down maps (BEV) images has become important for autonomous driving~\cite{saha2022translating,zhou2022cross,yang2021projecting}. Most works directly learn frame-by-frame mapping in a data-driven fashion without taking into account ego-motion~\cite{bertoni2019monoloco,zhang2021body}, \ie, they are only relative to the observer. Nishimura et al. proposed view birdification which is the task of reconstructing on-ground positions and trajectories of pedestrians and the observer, \ie, simultaneous decoupling of ego-motion from observed 2D motions in the FPV and anchoring of all motions in absolute coordinates~\cite{nishimura2021bmvc}. Our work differs from BEV transform and view birdification in three critical points. First, our goal is \emph{predictive} ego2top, not just a mapping between two views but predicting constituent positions in this view transform. Second, we aim at constructing a Pedestrian World Model not from the appearance, but from the movements, which results in a compact and efficient representation that can easily generalize. Third, unlike previous approaches which only perform deterministic mappings~\cite{saha2022translating}, we derive a probabilistic formulation that can handle uncertainty underlying object distances in the FPV. 

\section{Pedestrian World Model}
\label{sec:pwm}
Our goal is to derive an on-ground \emph{Pedestrian World Model} (PWM), an object-oriented abstraction of a crowd on the ground, from egocentric views captured in the crowd. PWM consists of one observer as an actor and pedestrians visible to that observer as objects. We derive an action-conditioned transition model of the environment consisting of these objects that can directly learn from in-environment 2D egocentric perception.

Consider a mobile robot equipped with a vision sensor immersed in a crowd consisting of people walking towards their own destinations while interacting with each other. A key characteristic of the PWM we aim to derive is that, unlike past object-oriented world models~\cite{lin2020improving}, the observer is the actor and is also part of the environment who interacts with other objects (pedestrians). In our running example, the mobile robot with a vision sensor is the observer and its ego-motion can be obtained by an IMU or other sensors including vision-based methods (\eg, SLAM~\cite{bescos2018dynaslam} and View Birdification~\cite{nishimura2021bmvc}). 

The observer location at timestep $\tau$ is given by the relative rotation $R(\theta_\tau) \in SO(2)$ and translation $\bm t_\tau \in \mathbb R^2$ on the ground plane from the previous timestep $\tau - 1$. The rotation $R(\theta_\tau)$ and $\bm t_\tau$ directly constitute the observer 
action $\bm a_\tau = \left[ \theta_\tau | \bm t_\tau \right ]^\top \in \mathbb R^3$.

At every timestep $\tau \in \{1, \dots, T\}$, the robot captures an image $I_\tau$ through which it observes the in-image states $\xset_\tau = \{\bm x^1, \dots, \bm x^{N_\tau}\}$ of $N_\tau$ people visible from the robot. Each state $\bm x^n = \left [u^n, v^n, \delta u^n, \delta v^n \right ]^\top \; (n = 1,\dots,N_\tau)$ consists of the 2D center position $\left [ u^n, v^n \right]^\top$ and its velocity $\left [ \delta u^n, \delta v^n \right ]^\top$ in $I_\tau$ calculated from the previous and subsequent frames. These states can be computed from the images with an off-the-shelf multi-object tracker~\cite{wang2019towards}, and the robot can keep track of $\{1,\dots, N_\tau\}$ pedestrians appearing across frames within a time window centered at $\tau$.

Given a set of in-image pedestrian states $\xset_\tau$ at current time step $\tau$ and the observer action $\bm a_\tau$, our objective is to predict future on-ground states of pedestrians $\yset_{\tau+1} = \{\bm y^1, \dots, \bm y^{N_\tau}\}$, where $\bm y^n = \left [ x^n, y^n, \delta x^n, \delta y^n \right ]^\top$. The on-ground pedestrian states $\bm y$ are described in the observer's camera coordinates, \ie, we predict 2D on-ground locations and their velocities relative to the observer's view which is then converted to absolute coordinates with the known observer state.
Our aim is to construct a transition model $\mathcal T(\xset_\tau| \bm a_\tau) \mapsto \yset_{\tau+1}$ that predicts future on-ground pedestrian states $\yset_{\tau+1}\in \mathbb R^{N_\tau \times 4}$ conditioned on the observer's action $\bm a_\tau$. The transition model can be formulated as

\begin{equation}
  \bm H_\tau = \mathcal V(\xset_{\tau})\,, \quad \yset_{\tau+1} = \mathcal G(\bm H_\tau, \bm Y_\tau, \bm a_\tau)\,,
  \label{eq:world-model}
\end{equation}
where $\mathcal V(\cdot)$ is the \emph{Vision Module} that encodes the ego-centric view observation $\xset_{\tau} \in \mathbb R^{N_\tau \times 4}$ into a $d$-dimensional embedding $\bm H_\tau \in \mathbb R^{N_\tau \times d}$ which represents a set of pedestrian state embeddings in an ego-centric view, and $\mathcal G(\cdot)$ is the \emph{Geometric Memory Module} which makes predictions of the future on-ground pedestrian states $\yset_{\tau+1}$ based on the past on-ground state estimates $\yset_{\tau}$ and the current observation $\xset_\tau$ with an implicit transform between the ego-centric view and the top-down ground view.

\section{\methodname}
\label{sec:incrowdformer}
\begin{figure}[t]
\centering
\includegraphics[width=\linewidth]{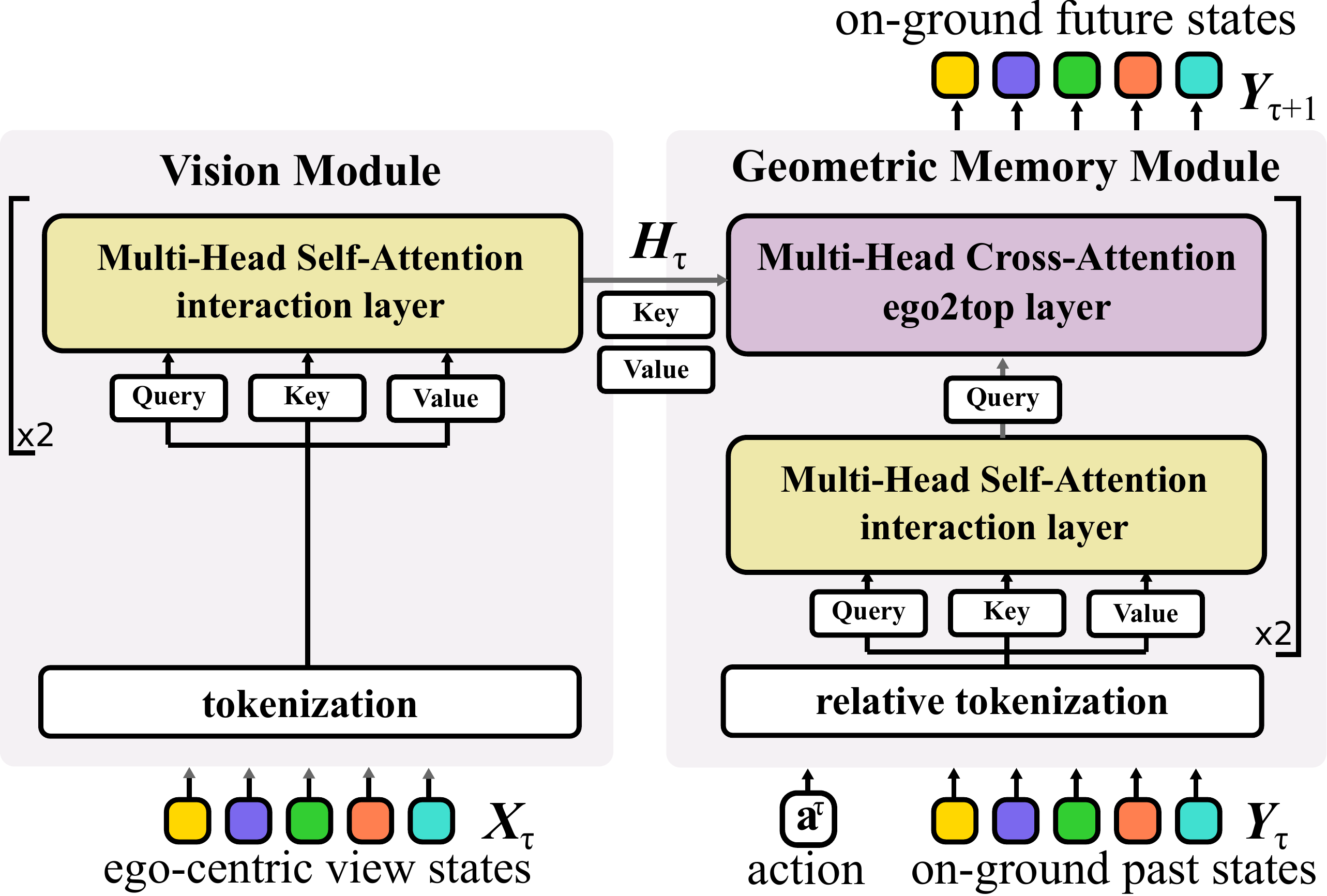}
\caption{\methodname consists of two modules referred to as the Vision Module and the Geometric Memory Module. The Vision Module encodes in-image interactions with self-attention to produce the current state embeddings $\bm H_\tau$ in an ego-centric view. The Geometric Memory Module predicts on-ground future states of pedestrians with cross-attention between the ego-centric view and on-ground past trajectories.}
\label{fig:overview}
\end{figure}

We introduce \methodname, a Transformer-based World Model, that realizes a PWM with object-centric interaction, ego2top transform, action conditioning, for a variable number of pedestrians. As \cref{fig:overview} depicts, \methodname has two modules, the Vision Module $\mathcal V$ and the Geometric Memory Module $\mathcal G$. 

\subsection{Vision Module}
The attention mechanism naturally provides a means to learn a mapping between two sets with variable numbers of constituents. We fully leverage this to learn the key ingredients, namely pedestrian interaction and ego2top transform both naturally conditioned on the observer actions, of an on-ground Pedestrian World Model from egocentric views.

The standard attention mechanism consists of a set of query vectors $\bm Q$, key vectors $\bm K$, and value vectors $\bm V$. These vectors are generated from the input tokens $\bm f_s$ and $\bm f_t$ as $\bm Q= W_Q\bm f_t\,,\quad \bm K = W_K\bm f_s\,,\quad \bm V = W_V\bm f_s$, where $W_Q$, $W_K$, and $W_V$ are linear embedding matrices. 
In \emph{self-attention} $\bm f_s = \bm f_t$ and for \emph{cross-attention} we have $\bm f_s \neq \bm f_t$. Unless otherwise noted, we use multi-head attention~\cite{vaswani2017attention} for both self-attention and cross-attention layers.

To model the interactions between pedestrians in the current frame, we use self-attention to learn the object-wise interactions observed in the ego-centric view during the state encoding process. As \cref{fig:overview} left depicts, the Vision Module $\mathcal V$ first tokenizes input first-person-view (FPV) state vectors into $d_s$-dimensional tokens with multi-layer perceptron (MLP) layers, $\xset_\tau \mapsto \bm f_s \in \mathbb R^{N_\tau \times d_s}$. It then computes self-attention over queries $\bm Q$ of tokens $\bm f_s$, where we encode their interactions in the image space into intermediate embeddings $\bm H_\tau$.

\subsection{Geometric Memory Module}
Learning a Pedestrian World Model is fundamentally different from standard trajectory forecasting problems~\cite{yuan2021agent} in that the on-ground movements of a crowd that needs to be predicted is deeply intertwined with the observer's ego-motion (\ie, action). That is, how the pedestrians move relative to the observer is largely affected by the observer's action. For this, we model a mapping  $\mathcal G(\bm a, \bm H_\tau, \yset_\tau) \mapsto \yset_{\tau+1}$ by $\mathcal G = \mathcal F_c(\mathcal F_s(\bm a_\tau, \yset_\tau), \bm H_\tau)$, where we use self-attention $\mathcal F_s$ to capture social interactions and cross-attention $\mathcal F_c$ to model the ego2top transform. This Geometric Memory Module first tokenizes the on-ground action and past pedestrian states as $\{\bm a_\tau, \yset_\tau\} \mapsto \bm f_t \in \mathbb R^{(N_\tau +1)\times d_s}$ with MLPs and computes self-attention over queries $\bm Q$ from tokens $\bm f_t$ to encode the interactions between the action and on-ground pedestrian trajectories. 
The cross-attention block takes queries $\bm Q \in \mathbb R^{(N_\tau+1) \times d_s}$ from the output of the self-attention layer and key, values $\bm K,\bm V \in \mathbb R^{N_\tau \times d_s}$ from the output of the vision module~(\Cref{fig:overview} left).

The geometric memory module is autoregressive, which means the module predicts future on-ground states one step at a time and uses the current prediction as input to make future predictions on a subsequent timestep. As the FPV view state embeddings $\bm H_\tau$ are updated by the vision module $\mathcal V$ at every timestep, their on-ground future predictions $\yset_{\tau+1}$ are generated by the geometric memory module $\mathcal G$ from the past predictions $\yset_\tau$, the current observer's action $\bm a_\tau$, and the FPV state embeddings $\bm H_\tau$.

We use two linear projections; $\varphi_p: \mathbb R^4 \mapsto \mathbb R^{d_s}$ for encoding in-image and on-ground pedestrian states, $\xset$ and $\yset$, and $\varphi_a : \mathbb R^3 \mapsto \mathbb R^{d_s}$ for the observer's action $\bm a$. We hard-concatenate positional information into states as $\bm x = \left [u, v \right ] \oplus \left [ \delta u, \delta v\right ]$. To encourage our model to generalize over diverse combinations of pedestrian tokens and observer actions, we transform on-ground pedestrian tokens relative to the observer's action token at every timestep.

\methodname can handle a varying number of pedestrians, \ie, tokens, in two ways. First, the attention matrix can be constructed with arbitrary sizes of input keys $\bm K$ and queries $\bm Q$. Second, an attention mask $M \in \mathbb R^{(N_\tau+1) \times N_\tau}$, where $M_{ij} = 0$ can be applied, if the pedestrian id $i$ is missing (\eg, occluded) at frame $\tau$, otherwise $1$.
The masked attention $\hat A$ becomes $\hat A_\tau = M_\tau \odot A_\tau$, where $\odot$ denotes the Hadamard product.

\section{Probabilistic \methodname}
\label{sec:cvae}

\begin{figure*}[t]
\centering
\includegraphics[width=0.9\linewidth]{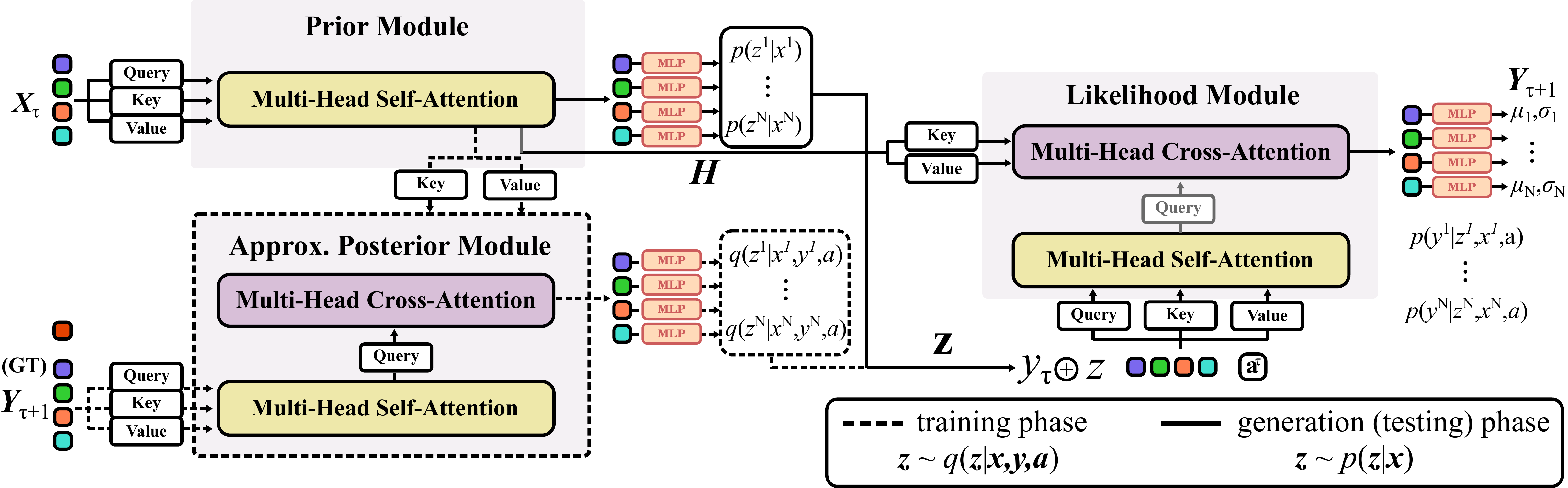}
\caption{Overview of Probabilistic \methodname. The output MLP layer of the Vision Module models the prior distribution. Each of the two geometric memory modules models the approximated posterior distribution and likelihood distribution, respectively. The latent codes are sampled from the approximated posterior module during training, and from the prior module at inference time.}
\label{fig:cvae}
\end{figure*}

We encode uncertainties arising from unknown pedestrian heights by making \methodname reason probabilistically on the pedestrian positions and their transitions. \Cref{fig:cvae} depicts an overview of the training and testing process of our Probabilistic \methodname.

Given a triplet of the current observed pedestrian states in the ego-centric view $\xset_\tau$, the current observer's action $\bm a_\tau$, and future pedestrian states  $\yset_{\tau+1}$, we construct a generative model based on Conditional VAE (CVAE). The CVAE learns the target on-ground pedestrian trajectory distributions in the future $p(\yset_{\tau+1} | \xset_\tau, \bm a_\tau)$ conditioned on the current ego-centric view observation $\bm X_\tau$ and the observer's action $\bm a_\tau$ through a stochastic latent variable $\bm Z$. Our model consists of a conditional prior network $p_\psi(\bm Z | \bm X_\tau)$ to model the latent variable $\bm Z$ from the observation $\bm X$, an approximated posterior network $q_\phi (\bm Z|\bm Y_{\tau+1},\bm X_\tau, \bm a_\tau )$, and a likelihood network $p_\eta (\bm Y_{\tau+1} | \bm Z, \bm X_{\tau}, \bm a_\tau )$. The objective of the CVAE model is to minimize the negative evidence lower bound (ELBO) in our loss function $\mathcal L$
\begin{align}
\thickmuskip=0mu
\medmuskip=0mu
\thinmuskip=0mu
 \begin{split}
  \label{eq:elbo}
     \mathcal L_{\text{ELBO}} =  &\mathbb E_{q_\phi (\bm Z|\bm Y, \bm X)} [ \log \underbrace{p_\eta (\yset_{\tau+1} | \bm Z, \xset_{\tau}, \bm a_\tau)}_{\text{likelihood}} ]\\
    &- \mathrm{KL} [ \underbrace{q_\phi (\bm Z|\yset_{\tau+1},\xset_\tau, \bm a_\tau)}_{\text{approximated posterior}}\| \underbrace{p_\psi(\bm Z | \xset_\tau)}_{\text{prior}} ]\,,
   \end{split}
\end{align}
where the first term maximizes the expectation of the log-likelihood of the target future state in the predicted distribution, and the second term minimizes the Kullback-Leibler (KL) divergence between the approximated posterior distribution and the prior distribution. Please see Supplementary Material A. for a more detailed derivation. In what follows, we describe how we implement each distribution in our Probabilistic \methodname.

\vspace{-12pt}
\paragraph{Prior Module} The \emph{Vision Module} can be seen as as a prior network that produces latent codes $\bm Z = \{\bm z^1, \dots, \bm z^{N_\tau}\}$, each of which follows a Gaussian distribution $p(\bm z^n | \xset_\tau) \sim  \mathcal N(\mu^n, (\sigma^n)^2)$. We process the output of the Vision Module $\bm H_\tau = \{\bm h^1, \dots, \bm h^{N_\tau}\}$ with a single MLP layer to map them into parameters of each Gaussian distribution ($\mu^n_p, \sigma^n_p)$.

\vspace{-12pt}
\paragraph{Approximated Posterior Module}
To approximate the posterior distribution $q_\phi (\bm z^n |\yset_{\tau+1},\xset_\tau,\bm a_\tau)$, we use a Transformer decoder with the same self-attention and cross-attention layers as the \emph{Geometric Memory Module}. The cross-attention layers in the Transformer decoder can efficiently model conditional relationships by taking the embeddings of FPV states $\bm H_\tau$ as keys and values, and the action and the ground-truth future state on the ground $\{\bm a_\tau, \bm Y_{\tau+1}\}$ as queries. We also assume the approximated posterior distribution follows a Gaussian distribution $q_\phi (\bm z^n | \yset_{\tau+1}, \xset_\tau, \bm a_\tau) \sim \mathcal N(\mu^n, (\sigma^n)^2)$. MLP layers process the output of the Transformer Decoder to obtain the Gaussian parameters $(\mu^n_q, \sigma^n_q)$ as in the Prior Module. Note that this approximated posterior module is used only during training.

\vspace{-12pt}
\paragraph{Likelihood Module}
We can view the \emph{Geometric Memory Module} as a likelihood module to model $p_\eta(\yset_{\tau+1} | \bm Z , \xset_\tau, \bm a_\tau)$. The latent codes $\bm Z$ for pedestrians are computed from the approximated posterior module during training, and from the vision module for inference. As we obtain predictions of future pedestrian states autoregressively from the geometric memory module, we feed $\yset$ concatenated with corresponding latent codes $\bm Z$ alongside the observer's action $\bm a_\tau$. We concatenate $\bm y^n \oplus \bm z^n$ for each pedestrian and tokenize these as input queries. The cross-attention layers in the geometric memory module effectively models the conditional relationship between $\bm Z$ and $\bm a_\tau$ and outputs a likelihood distribution. We consider two variants of the proposed method based on the definition of the outputs, \textbf{InCrowdFormer-D} and \textbf{InCrowdFormer-G}.

InCrowdFormer-D directly estimates the future pedestrian states and, for the first term of \cref{eq:elbo}, uses the mean squared error (MSE), $\mathcal L_\text{D} = \| \hat {\yset} - \yset \|^2$, where $\hat{\yset}$ is the ground-truth future state of pedestrians on the ground.

InCrowdFormer-G takes into account uncertainties arising from ego2top transformation of unknown object scales as aleatoric uncertainty~\cite{kendall2017uncertainties} computed as a probability distribution over the \methodname outputs. We model the uncertainty with a 2D Gaussian distribution on the ground $p(\yset_{\tau+1} | \bm Z, \bm X_\tau, \bm a_\tau)\sim \mathcal N(\bm \mu^n, \Sigma^n)$, and make \methodname output these Gaussian parameters for each pedestrian by defining the first term of \cref{eq:elbo} as a negative log-likelihood loss of a 2D Gaussian function $\mathcal L_{\text{G}} =  \frac{\| \hat{\yset_x} - \yset_x  \|^2}{(2\sigma_x)^2} + \frac{\| \hat{\yset_y} - \yset_y  \|^2}{(2\sigma_y)^2}$.
Such a parametric representation of uncertainty of the future pedestrian state would be useful for downstream applications. 

\section{Experiments}
We evaluate the effectiveness of Probabilistic \methodname (hereafter simply \methodname) trained on real-world crowd trajectories with augmented observer's actions in terms of its $T$-step prediction accuracy and also demonstrate its application to real video sequences. We provide more implementation details in the supplementary material.

\begin{table*}[t]
  \caption{Quantitative Results of \methodname applied to the ego-centric view crowd dataset. (IV) and (CV) indicate the intra-scene, and the cross-scene validation split, respectively. \metric{ADE}{T} and \metric{FDE}{T} represent average and final displacement errors [m] in $T$-step prediction. \methodname variants clearly outperform baseline methods in terms of $5$-step and $10$-step prediction accuracy. Prediction results applied to the cross-scene validation split dataset demonstrate that our method can be generalized to unknown crowd scenarios.} 
  \vspace{-1.5\baselineskip}  
  \begingroup
  \scriptsize
  \begin{center}
  \renewcommand{\arraystretch}{0.9}
  \begin{tabularx}{0.9\linewidth}{l|cccc|cccc|cccc}
    \toprule[1.5pt]
    Dataset (IV) & \multicolumn{4}{c}{Hotel (sparse)}  & \multicolumn{4}{c}{ETH (mid) } & \multicolumn{4}{c}{Students (dense)}\\
    \toprule[1pt]
     Metrics & \metric{ADE}{5} & \metric{FDE}{5} & \metric{ADE}{10} & \metric{FDE}{10} & \metric{ADE}{5} & \metric{FDE}{5} & \metric{ADE}{10} & \metric{FDE}{10} & \metric{ADE}{5} & \metric{FDE}{5} & \metric{ADE}{10} & \metric{FDE}{10} \\
     \midrule
     MLP-TSSM~\cite{Dugas2022navdreams,chen2022transdreamer} & 4.009 & 4.763 & 6.014 & 6.432 & 4.102 & 4.835 & 5.986 & 6.287 & 5.939 & 6.126 & 6.723 & 6.921 \\
     MLP-RSSM~\cite{ha2018worldmodels} & 4.251 & 4.861 & 6.027 & 6.853 & 4.562 & 4.925 & 6.102 & 6.489 & 6.102 & 6.615 & 6.532 & 6.911 \\
     C-SWM~\cite{kipf2019contrastive} & 1.475 & 1.592 & 1.972 & 2.018 &  1.535 & 1.682 & 1.920 & 2.012 & 1.428 & 1.492 & 2.156 & 2.215 \\ 
     SlotFormer~\cite{wu2022slotformer} & 1.512 & 1.618 & 1.815 & 1.901 & 1.631 & 1.721 & 1.810 & 1.912 & 1.419 & 1.521 & 2.016 & 2.161 \\ 
    \midrule
    \rowcolor[rgb]{0.93,1.0,0.87}{\textbf{\scalebox{0.9}[1]{InCrowdFormer-D}}}& 0.358 & 0.468 & 0.626 & 0.829 & 0.326 & 0.385 & 0.537 & 0.683 & 0.251 & 0.273 & 0.432 & 0.483 \\
    \rowcolor[rgb]{0.93,1.0,0.87}{\textbf{\scalebox{0.9}[1]{InCrowdFormer-G}}}& 0.327 & 0.472 & 0.648 & 0.818 & 0.372 & 0.393 & 0.526 & 0.652 & 0.312 & 0.372 & 0.461 & 0.492 \\
    \bottomrule
        \toprule[1pt]
    Dataset (CV) & \multicolumn{4}{c}{Hotel (sparse)}  & \multicolumn{4}{c}{ETH (mid) } & \multicolumn{4}{c}{Students (dense)}\\
    \toprule[1pt]
     MLP-TSSM~\cite{Dugas2022navdreams,chen2022transdreamer} &4.228 &4.934 &6.100 &6.353 &4.147 &4.941 &6.175 &6.318 &6.064 &6.208 &6.737 &7.007 \\
     MLP-RSSM~\cite{ha2018worldmodels}& 4.315 &4.856 &6.223 &6.843 &4.549 &5.020 &6.105 &6.615 &5.849 &6.649 &6.544 &7.002 \\
     C-SWM~\cite{kipf2019contrastive} & 1.515 & 1.613 & 2.215 & 2.251 &  1.721 & 1.810 & 2.018 & 2.231 & 1.615 & 1.712 & 2.381 & 2.519 \\ 
     SlotFormer~\cite{wu2022slotformer} & 1.618 & 1.712 & 1.920 & 2.060 & 1.809 & 1.910 & 1.915 & 2.016 & 1.591 & 1.612 & 2.130 & 2.210 \\ 
    \midrule
    \rowcolor[rgb]{0.93,1.0,0.87}{\textbf{\scalebox{0.9}[1]{InCrowdFormer-D}}} & 0.687 &0.638 &0.717 &1.037 &0.469 &0.421 &0.580 &0.601 &0.323 &0.383 &0.512 &0.568 \\
    \rowcolor[rgb]{0.93,1.0,0.87}{\textbf{\scalebox{0.9}[1]{InCrowdFormer-G}} }& 0.536 &0.272 &0.714 &0.881 &0.372 &0.428 &0.646 &0.722 &0.437 &0.523 &0.578 &0.720 \\
    \bottomrule[1.5pt]
  \end{tabularx}
  \end{center}
  \endgroup
  \label{table:quantitative-results}
\end{table*}

\vspace{-12pt}
\paragraph{Ego-centric View Crowd Dataset}
We construct a crowd dataset consisting of on-ground real pedestrian trajectories paired with their state features in an ego-centric view. We first extract trajectories referred to as \textbf{Hotel}, \textbf{ETH}, and \textbf{Students} from the ETH~\cite{pellegrini2009you} and the UCY~\cite{lerner2007crowds} datasets. These three sets of trajectories correspond to sparse, moderate, and dense crowds. To compute the egocentric views of those pedestrians, we first sample pedestrian heights from a Gaussian distribution $\mathcal N(\mu, \sigma)$ with $\mu = 1.70$ m and $\sigma = 0.07$ m according to the statistics of European adults~\cite{visscher2008}. The virtual head points with the height $h \sim \mathcal N(\mu, \sigma)$ are then projected to the 2D positions $\left [u, v \right]$ by perspective projection with a known intrinsic matrix $A$. For each pedestrian, we compute the state feature $\left [u, v, \delta u, \delta v\right ]$ in the ego-centric view and pair it with its corresponding on-ground position. Since our proposed Pedestrian World Model is an abstraction of the on-ground crowd movements, we do not need photo-realistic renderings in the datasets. This allows us to augment datasets with diverse combinations of ego-motion and crowd pedestrian trajectories easily and efficiently, which is otherwise challenging, if not impossible, to collect in the real world.

\vspace{-12pt}
\paragraph{Observer Action Generation}
To model the transition of the world in the observer's view when navigating in a crowd while avoiding potential collisions, we generate plausible observer's trajectories with an Optimal Reciprocal Collision Avoidance (ORCA) planner~\cite{van2011reciprocal} in the ego-centric view crowd dataset. We randomly sample starting positions on a circle with a fixed radius $r=8.0$ m and the observer walks to its destination set at the opposite side of the circle by the planner. We mount two virtual, perspective cameras in front and rear on the observer, each of which captures states of the pedestrians in the crowd.

\vspace{-12pt}
\paragraph{Baselines}
Our method is the first to realize an object-oriented, on-ground world model conditioned on the observer's action trajectory solely from ego-centric in-crowd observations.
Note that there is no existing method that achieves exactly the same task. We compare with original, object-oriented baselines composed of the cascaded object encoder and the memory module which are components of the state-of-the-art world models~\cite{kipf2019contrastive,wu2022slotformer,ha2018worldmodels,chen2022transdreamer}. In particular, we use a standard MLP object encoder/decoder and memory modules referred to as Recurrent State Space Model (RSSM)~\cite{ha2018worldmodels} and Transformer State Space Model (TSSM)~\cite{chen2022transdreamer}. We refer to these baselines as \textbf{MLP-RSSM} and \textbf{MLP-TSSM}. We also compare our method with variants of recent object-oriented world models. We extend \textbf{C-SWM}~\cite{kipf2019contrastive} and \textbf{SlotFormer}~\cite{wu2022slotformer} to consider the observer's action in their memory module and combine it with the standard MLP object encoder/decoder. In their original implementations, these world models encode transitions in ego-centric views, unlike our model which encodes on-ground transitions conditioned on the ego-views. Furhter, we consider two variants of \methodname, \textbf{InCrowdFormer-D} (deterministic) and \textbf{InCrowdFormer-G} (Gaussian) as introduced in \cref{sec:cvae}. 

\begin{figure*}
    \centering
    \includegraphics[width=\linewidth]{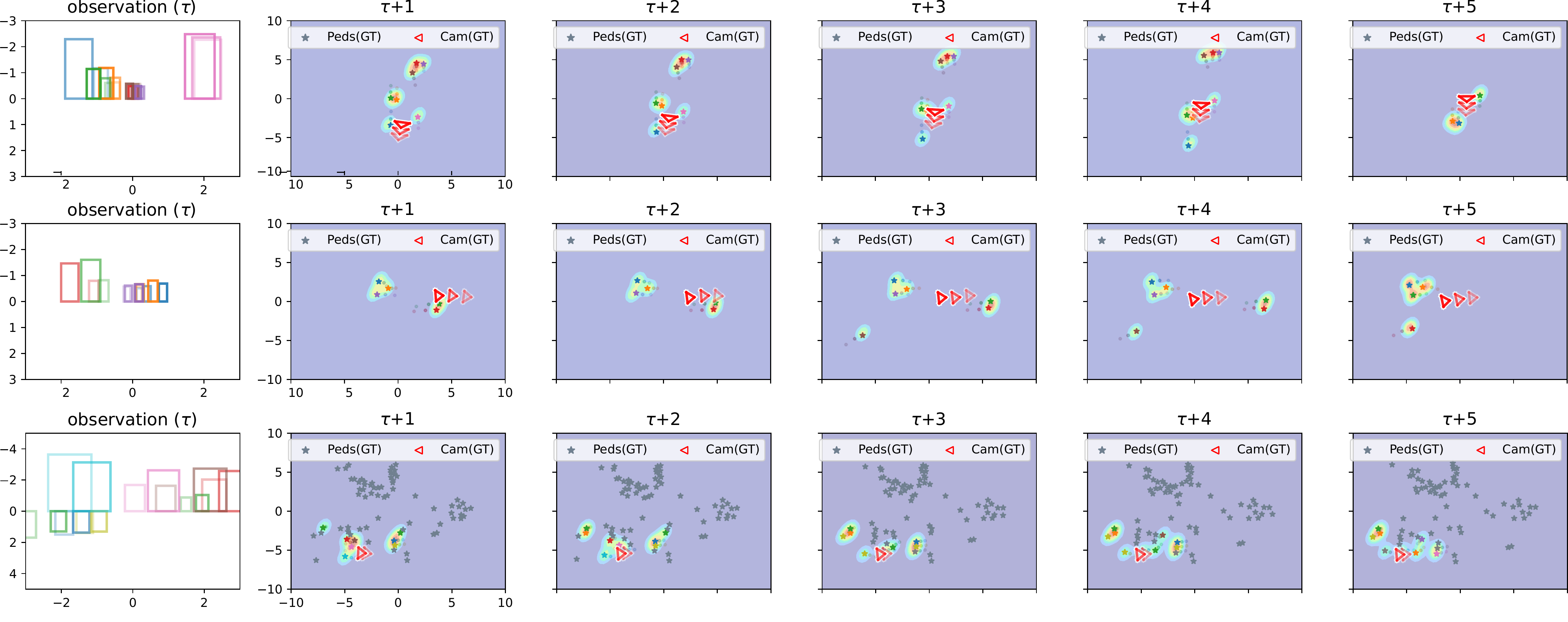}
    \caption{Qualitative results of InCrowdFormer-G applied to the Hotel (top row), ETH (middle), and Students (bottom) datasets. Left most: First frames in the ego-centric views. Rest: On-ground future prediction results in subsequent frames. Negative heights of bounding boxes depict observations from the rear camera. The probabilities are rendered with red (high) to blue (low) heatmaps. Stars depict the ground-truth positions. Our model successfully predicts on-ground pedestrian states for crowds of diverse densities. Even in a dense crowd, our method can successfully handle the varying number of pedestrians and predict accurate future locations of nearby pedestrians.}
    \label{fig:qualitative-results}
\end{figure*}

\begin{figure*}
    \centering
    \includegraphics[width=\linewidth]{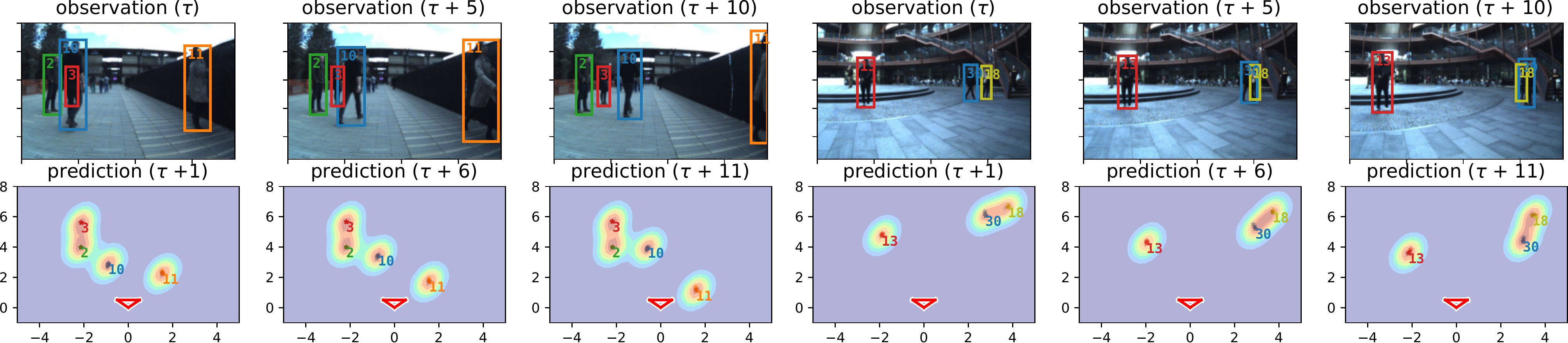}
    \caption{Inference results of InCrowdFormer-G applied to real video sequences from the JRDB dataset. Top row shows input ego-centric view video and pedestrian bounding boxes. Bottom row shows on-ground $\tau+1$ future pedestrian trajectories predicted by our model. Our method can easily be adapted to real video sequences as it only requires the positions and scales of the bounding boxes of pedestrians.}
    \label{fig:jrdb}
\end{figure*}

\vspace{-12pt}
\paragraph{Metric} 
Given a sequence of observer's actions and corresponding ego-centric view observations, we predict the on-ground state of pedestrians in the future by autoregressively feeding an action $\bm a_\tau$ and ego-centric view states $\xset_\tau$ into our \methodname model. For a set of action sequences $\{\bm a_1, \dots, \bm a_T \}$ and corresponding ego-centric view states $\{\xset_1, \dots, \xset_T \}$ during $T$-steps in the testing set, we evaluate the prediction accuracy with Average Displacement Error (ADE) and Final Displacement Error (FDE). We define \metric{ADE}{T} $= \frac{1}{T N^T} \sum_{n=1}^{N^T} \sum_{\tau=1}^T \| \hat{\bm y}^n_\tau - \bm y^n_\tau \|$ and \metric{FDE}{T} $= \frac{1}{N^T}\sum_{n=1}^{N^T} \|\hat{\bm y_T^n} - \bm y_T^n \|$, respectively. For InCrowdFormer-D and InCrowdFormer-G, we compute the best-of-10 results, \ie, the minimum ADE and FDE from 10 randomly-sampled latent codes as in~\cite{gupta2018social}.

\subsection{Quantitative evaluation of prediction accuracy}
We prepare two validation splits for the ego-centric view crowd dataset. One is the intra-scene validation split, and the other is the cross-scene validation split, \ie, leave-one-out cross-validation. \Cref{table:quantitative-results} shows the prediction accuracy evaluated in \metric{ADE}{T} and \metric{FDE}{T} with $T \in \{5,10\}$.  The RSSM baseline makes a prediction without taking object interactions into account, which results in the lowest accuracy in all the metrics. 
Both the InCrowdFormer-D and InCrowdFormer-C outperform all the baselines in terms of short-term ($T=5$) and long-term ($T=10$) prediction accuracy. 
Finally, our method achieves high accuracy on the dataset with the cross-scene validation split. This demonstrates that our method can be generalized to unknown scenes by training the model with a diverse density of crowds. Most important, our method still achieves high ($<1.0$ [m]) accuracy for long-term ($T=10$) prediction sequences as well as short-term ($T=5$) sequences by attending to observed ego-centric view features at every timestep.

\vspace{-12pt}
\paragraph{Qualitative Results}
\Cref{fig:qualitative-results} visualizes the prediction results of InCrowdFormer-G from time $\tau$ to $\tau+5$ frames on the Hotel, ETH, and Students datasets. InCrowdFormer-G outputs the future location of a pedestrian with a 2D Gaussian distribution which captures uncertainty arising from imperfect cues of depths and pedestrian interactions. 
Our method also predicts future locations of nearby pedestrians in a dense crowd by masking the attention matrix for occluded pedestrians.
These outputs are beneficial to downstream applications such as robot crowd navigation, where we should path-plan to avoid potential collisions with nearby pedestrians from an ego-centric view and limited depth information. 

\subsection{Ablation Study}
\paragraph{Effect of ego-centric view for prediction}
To evaluate the effectiveness of the cross-attention between  the ego-centric and the on-ground view inputs, we prepare a variant of our method that takes only on-ground past features as inputs and outputs on-ground future trajectories by self-attention layers. The first and the third row of \Cref{tab:ablation} clearly demonstrate that our ego-centric view-conditioned future prediction outperforms on-ground future trajectory prediction that depends only on the on-ground past trajectories.

\vspace{-12pt}
\paragraph{Effect of cross-attention}
We also prepare a variant of our method consisting of only self-attention layers. For this variant, we replace the geometric memory module with one that consists of just two self-attention layers. The modified geometric memory module takes the output of the vision module as inputs and calculates the on-ground future states with self-attention. In \Cref{tab:ablation}, the variants of our model without the cross-attention (second row) underperform in terms of prediction accuracy, which clearly demonstrates that the cross-attention between the vision module and the geometric memory module is essential to our model.

\vspace{-12pt}
\paragraph{Effect of uncertainty modeling}
We compare two variants of the proposed method based on the choice of the outputs. In \Cref{table:quantitative-results}, overall, InCrowdFormer-D achieves slightly higher accuracy than InCrowdFormer-G, which can be attributed to the fact that InCrowdFormer-D directly regresses future positions while InCrowdFormer-G predicts uncertainty distributions.

\begin{table}[t]
  \begingroup
  \tabcolsep = 1.5mm
  \scriptsize
    \caption{Ablation study of the cross-attention mechanism in our proposed method. These results clearly demonstrate that the cross-attention between the vision module and the geometric memory module is essential in our model.}
          \label{tab:ablation}
        \vspace{-6pt}
  \begin{tabularx}{\linewidth}{cc|cccccc}
    \toprule[1.2pt]
    \multicolumn{2}{c}{\textbf{Dataset (IV) }} & \multicolumn{2}{c}{\textbf{Hotel (sparse)}} & \multicolumn{2}{c}{\textbf{ETH (mid)}} & \multicolumn{2}{c}{\textbf{Students (dense)}} \\
    \toprule[1pt]
     ego-view & cross-attn & \metric{ADE}{5} & \metric{ADE}{10} & \metric{ADE}{5} & \metric{ADE}{10} & \metric{ADE}{5} & \metric{ADE}{10} \\
    \toprule[1pt]
     -- & --& 0.482 & 0.648 & 0.495 & 0.482 & 0.391 & 0.523 \\
     \checkmark & --& 0.526 & 0.702 & 0.546 & 0.728 & 0.451 & 0.826 \\
    \midrule
       \rowcolor[rgb]{0.93,1.0,0.87}  \checkmark & \checkmark & \textbf{0.358} & \textbf{0.626} & \textbf{0.326} & \textbf{0.537} & \textbf{0.251} & \textbf{0.432} \\
    \bottomrule[1.2pt]
  \end{tabularx}
  \endgroup
\end{table}

\subsection{Inference on Real Data}
\Cref{fig:jrdb} shows the inference results of InCrowdFormer-G applied to real video sequences from the JRDB dataset~\cite{martin2021jrdb}. We first train our model with the same camera parameter of JRDB. We train on the ETH dataset, which is similar in crowd density to the JRDB sequences. We then apply our pre-trained model to the target sequences. Our method can easily be adapted to real video sequences as it requires only the position and scale of pedestrian bounding boxes and abstracts away their appearance. 

\section{Conclusion}
In this paper, we introduced \methodname, a Transformer-based Pedestrian World Model that predicts on-ground pedestrian trajectories from an ego-centric view observation. Our extensive evaluation demonstrates the effectiveness of our approach in diverse density of crowds, and also show promising results in zero-shot adaptation to real video sequences. We believe our \methodname will serve as a sound foundation for crowd and pedestrian movement modeling and enable a wide range of downstream applications including but not limited to navigation. We will release all the code and data to catalyze such use.

\vspace{-14pt}
\paragraph{Limitation} In this study, we assumed perfect odometry observation for the observer's action, \ie, known observer ego-motion, and did not consider observation noise in the ego-centric views. We believe that this can be modeled similarly as uncertainty in depth arising from pedestrian's height variances, which we plan to explore in future work.

\appendix
\section{Conditional VAE (CVAE)}
We provide further details of our Probabilistic InCrowdFormer described in Section 5. of the main manuscript.
To model the distribution of the future on-ground states of pedestrians $p(\yset_{\tau+1}| \xset_\tau, \bm a_\tau)$ conditioned on the ego-centric view observation $\xset_\tau$ and the observer's action $\bm a_\tau$, we formulate it as a Conditional Variational Autoencoder (CVAE)~\cite{sohn2015learning}. 
CVAE is a conditional generative model that outputs the target $\bm Y$ conditioned on the latent variable $\bm Z$ and the input observation $\bm X$.
We introduce $d_z$-dimensional latent codes for each pedestrian $\bm Z = \{\bm z_1, \dots, \bm z_{N_\tau}\} \in \mathbb R^{N_\tau \times d_z}$ and re-formulate the future distribution of the pedestrian states $\bm Y_{\tau+1}$ as
\begin{align}
\thickmuskip=0mu
\medmuskip=0mu
\thinmuskip=0mu
\label{eq:generative-model}
        p(\yset_{\tau+1} | \xset_{\tau}, \bm a_\tau) =\int \underbrace{p (\yset_{\tau+1} | \bm Z, \xset_\tau, \bm a_\tau)}_{\text{likelihood}} \underbrace{p(\bm Z |\xset_\tau)}_{\text{prior}} d\bm Z\,,
\end{align}
where $p(\yset_{\tau+1} | \bm Z, \bm X_\tau, \bm a_\tau)$ is the conditional likelihood and $p(\bm Z | \bm X_\tau) = \prod_n p(\bm z^n | \xset_\tau)$ is the conditional Gaussian prior factorized over pedestrians. The observer's action is deterministic in our Pedestrian World Model, so that latent codes are only necessary for pedestrians and not the observer.

Due to the intractable integral computation in \cref{eq:generative-model}, we instead minimize the negative evidence lower bound (ELBO) in our loss function $\mathcal L$
\begin{align}
  \begin{split}
  \label{eq:elbo}
        \mathcal L_{\text{ELBO}}(\xset_\tau, &\yset_{\tau+1}, \bm a_\tau;\eta, \phi, \psi) =\\
     &\mathbb E_{q_\phi (\bm Z|\bm Y, \bm X)} [ \log \underbrace{p_\eta (\yset_{\tau+1} | \bm Z, \xset_{\tau}, \bm a_\tau)}_{\text{likelihood}} ]\\
    &- \mathrm{KL} [ \underbrace{q_\phi (\bm Z|\yset_{\tau+1},\xset_\tau, \bm a_\tau)}_{\text{approximated posterior}}\| \underbrace{p_\psi(\bm Z | \xset_\tau)}_{\text{prior}} ]\,,
  \end{split}
\end{align}
where the first term maximizes the expectation of the log-likelihood of the target future state in the predicted distribution, and the second term minimizes the Kullback-Leibler (KL) divergence between the approximated posterior distribution and the prior distribution.

Consequently, our CVAE-based probabilistic InCrowdFormer consists of the three modules, a conditional prior network $p_\psi(\bm Z|\bm X_\tau)$, an approximated posterior network $p_\phi(\bm Z|\bm Y_{\tau+1},\bm X_\tau, \bm a_\tau)$, and a likelihood network $p_\eta (\bm Y_{\tau+1} | \bm Z, \bm X_\tau, \bm a_\tau).$
Given the current observed pedestrian states in the ego-centric view $\xset_\tau$, the observer's action $\bm a_\tau$, and future pedestrian state relative to the observer $\yset_{\tau+1}$, our goal is to learn a generative model $p(\yset_{\tau+1} | \xset_\tau, \bm a_\tau)$ by learning network parameters $\phi, \psi$, and $\eta$ by minimizing \cref{eq:elbo}. 

\section{In-Depth Analysis of Qualitative Results}
\begin{figure}
    \centering
    \includegraphics[width=\linewidth]{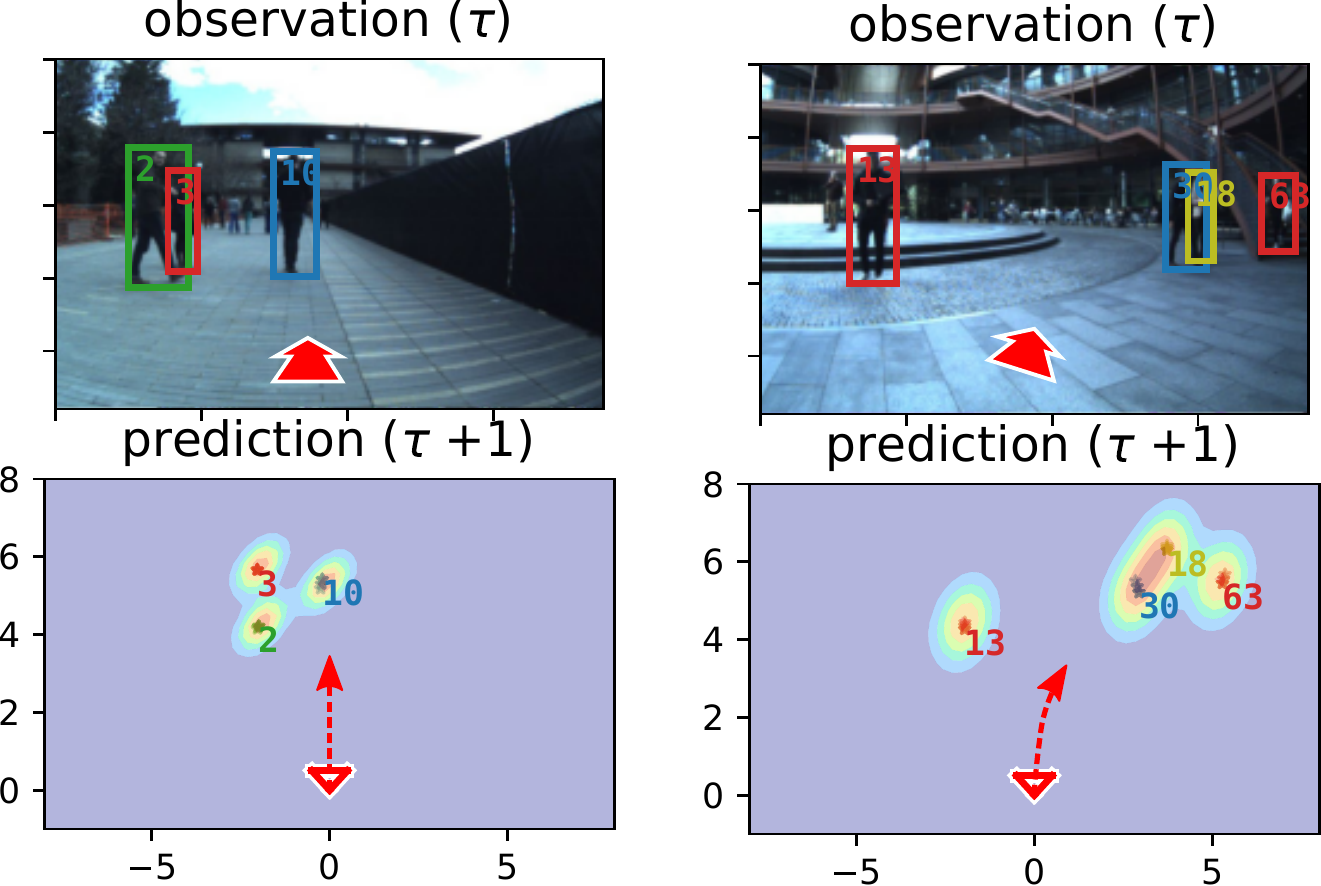}
    \caption{Prediction result of InCrowdFormer-G conditioned on a monotonous straightforward ego-motion (Left) and a curving ego-motion (Right). The variance of the Gaussian distribution corresponds to the uncertainty of the estimated location.}
    \label{fig:uncertainty}
\end{figure}
\Cref{fig:uncertainty} compares prediction results with respect to the observer's ego-motion. In the left case, both the camera and pedestrians are going straight and our model can easily localize the future locations of pedestrians certainly. In the right case, the predicted Gaussian distributions have heavier tails as the model needs to account for complex interactions between the observer and the surrounding pedestrians. We expect that these probabilistic representations are indeed beneficial for downstream applications such as navigating in a crowd, where the observer seeks a path while avoiding potential collisions with nearby pedestrians.

\section{Additional Experimental Details}
We use an MLP with two layers and $16$ hidden dimension to encode the input state $\bm x$
into $d_s=32$ dimensional input tokens. The latent dimensions of $\bm z$ is set to $32$.
We use two self-attention layers in the vision module and also two self-attention and cross-attention layers in the geometric memory module. We set the number of heads for multi-head attention (MHA) to six in each attention layer.
For RSSM and TSSM, we set the input embedding size to $32$, which is the same size as $d_s$ of our method.
We also use another MLP with the same structure to encode and decode future state $\bm y$.
All of these models are implemented in PyTorch and trained with the Adam optimizer~\cite{kingma2014adam} for $800$ epochs with minibatch size $32$ and learning rate $0.001$. All experiments are run on a PC with Intel(R) Xeon(R) Gold 6752 CPU and an NVIDIA V100 GPU.

\section{Statistics of Crowd Dataset}
\Cref{tab:stats} shows the crowd density of each dataset. Based on the average number of people appearing in one frame, we refer to Hotel, ETH, and Students as sparse, moderate, and dense crowd scenarios, respectively. In each scenario, we generated $500$ trajectories of the moving observer with random starts and goals with ORCA~\cite{van2011reciprocal} for the training split and $200$ for the testing split.

\begin{table}[t]
  \caption{Crowd densities in each dataset. TrajLen denotes the average duration of each pedestrian's trajectory. $N$ People in Crowd shows the average, minimum, and maximum numbers of people appearing in a frame.}
  \begin{tabularx}{\linewidth}{l|cccc}
  \toprule[1.5pt] 
   & \textbf{TrajLen} & \multicolumn{3}{c}{\textbf{$N$ People in Crowd}}\\
Dataset & Avg & Avg & Min & Max \\
  \midrule[1pt]
    \textbf{Hotel} (sparse) & 15.0 & 6.31 &  3 & 15\\ 
    \textbf{ETH} (moderate) & 14.4 & 9.29 & 3 & 26 \\ 
    \textbf{Students} (dense) & 45.8 & 44.2 & 13 & 75\\
  \bottomrule[1.5pt]
  \end{tabularx}
  \label{tab:stats}
  \vspace{-12pt}
\end{table}

\section{Quantitative Evaluation on JRDB sequence}
JackRabbot Dataset and Benchmark (JRDB)~\cite{martin2021jrdb} includes RGB images with 2D-3D bounding box annotations for pedestrians captured by the JackRabbot moving platform.  We select two sequences where the moving platform is smoothly navigating in a crowd.  We refer to these two sequences, \emph{clark-center-intersection-2019-02-08\_0} and \emph{clark-center-2019-02-08\_0} as \textbf{clark-center I} and \textbf{clark-center II}. They correspond to Figure 5 left and Figure 5 right in the main text.
As described in the main text, we applied our InCrowdFormer-D trained with the ETH dataset.  We used the original camera parameter of JRDB for these two sequences and evaluated in \metric{ADE}{T} and \metric{FDE}{T}. Since JRDB does not provide the ground-truth of the ego-motion of the mobile platform, we fed the IMU values as the observer's action, which was parsed from their provided rosbag data. \Cref{tab:jrdb} shows quantitative results of InCrowdFormer-G applied to these two JRDB sequences. Even without re-training on the JRDB data, our method shows good prediction accuracy on the real video sequences by using only pedestrian bounding boxes extracted from the ego-centric images. 

\begin{table}[t]
  \caption{Quantitative results of InCrowdFormer-G applied to JRDB sequences. Our model trained with ETH dataset demonstrates good prediction accuracy.}
  \begin{tabularx}{\linewidth}{l|cccc}
  \toprule[1.5pt] 
  Metric & \metric{ADE}{5} & \metric{FDE}{5} & \metric{ADE}{10} & \metric{FDE}{10} \\
  \midrule[1pt]
    clark-center I & 0.325 & 0.386 & 0.421 & 0.459 \\ 
    clark-center II & 0.227 & 0.244 & 0.315 & 0.322 \\
\midrule[1pt]
    ETH (trained) & 0.372 & 0.393 & 0.526 & 0.652 \\
  \bottomrule[1.5pt]
  \end{tabularx}
  \label{tab:jrdb}
\end{table}

\balance
{\small
\bibliographystyle{ieee_fullname}
\bibliography{egbib}
}

\end{document}